\documentclass[letterpaper]{article} 
\usepackage{aaai23}  
\usepackage{times}  
\usepackage{helvet}  
\usepackage{courier}  
\usepackage[hyphens]{url}  
\usepackage{graphicx} 
\usepackage{natbib}  
\usepackage{caption} 
\usepackage{multirow}
\usepackage{booktabs}
\usepackage{amssymb}
\usepackage{amsmath}
\usepackage{amsthm}
\usepackage{algorithm}
\usepackage{algorithmic}

\usepackage{newfloat}
\usepackage{listings}
\DeclareCaptionStyle{ruled}{labelfont=normalfont,labelsep=colon,strut=off} 
\floatstyle{ruled}
\newfloat{listing}{tb}{lst}{}
\floatname{listing}{Listing}
\title{MobileTL: On-Device Transfer Learning with Inverted Residual Blocks}
\author {
    Hung-Yueh Chiang,
    Natalia Frumkin,
    Feng Liang,
    Diana Marculescu
}
\affiliations {
    The University of Texas at Austin \\
    Chandra Family Department of Electrical and Computer Engineering\\
    \{hungyueh.chiang, nfrumkin, jeffliang, dianam\}@utexas.edu
}

\usepackage{bibentry}
\newtheorem{theorem}{Theorem}
\newtheorem{lemma}{Lemma}
\newtheorem{proposition}{Proposition}

\makeatletter
\newtheorem{repeatthm@}{Theorem}
\newenvironment{repeatthm}[1]{%
    \def\therepeatthm@{\ref{#1}}
    \repeatthm@
}
{\endrepeatthm@}
\makeatother

\newcommand{\raisedtarget}[1]{%
  \raisebox{\fontcharht\font`P}[0pt][0pt]{\hypertarget{#1}{}}%
}

\begin{document}

\maketitle

\begin{abstract} 
Transfer learning on edge is challenging due to on-device limited resources.
Existing work addresses this issue by training a subset of parameters or adding model patches.
Developed with inference in mind, Inverted Residual Blocks (IRBs) split a convolutional layer into depthwise and pointwise convolutions, leading to more stacking layers, \textit{e.g.}, convolution, normalization, and activation layers.
Though they are efficient for inference, IRBs require that additional activation maps are stored in memory for training weights for convolution layers and scales for normalization layers.
As a result, their high memory cost prohibits training IRBs on resource-limited edge devices, and making them unsuitable in the context of transfer learning.
To address this issue, we present MobileTL, a memory and computationally efficient on-device transfer learning method for models built with IRBs.
MobileTL trains the shifts for internal normalization layers to avoid storing activation maps for the backward pass.
Also, MobileTL approximates the backward computation of the activation layer (\emph{e.g.}, Hard-Swish and ReLU6) as a signed function which enables storing a binary mask instead of activation maps for the backward pass.
MobileTL fine-tunes a few top blocks (close to output) rather than propagating the gradient through the whole network to reduce the computation cost.
Our method reduces memory usage by 46\% and 53\% for MobileNetV2 and V3 IRBs, respectively.
For MobileNetV3, we observe a 36\% reduction in floating-point operations (FLOPs) when fine-tuning 5 blocks, while only incurring a 0.6\% accuracy reduction on CIFAR10.
Extensive experiments on multiple datasets demonstrate that our method is Pareto-optimal (best accuracy under given hardware constraints) compared to prior work in transfer learning for edge devices.
\end{abstract}

\section{Introduction}
\label{introduction}
With the plethora of mobile devices available for consumers today, there is a demand for fast, customized, and privacy-aware deep learning algorithms.
To reach satisfactory performance, current deep learning trends promote models with billions of parameters \cite{dosovitskiy2021image, brown2020language, radford2021learning} whose on-device training is infeasible.
One solution is to train the model with data in the cloud and rely on efficient communication to update the model on the mobile device \cite{samie2016computation}.
However, applications with tight data privacy constraints often make sending data from mobile devices to the cloud infeasible and suffer from degraded performance \cite{liu2021machine}.
In this setting, we consider \textit{on-device transfer learning}, where a pre-trained model is downloaded to the mobile device and fine-tuned using local data.
This way, the model is properly adapted to the target domain without sending potentially sensitive data to central servers.

\begin{figure}
\centering
\includegraphics[width=75mm]{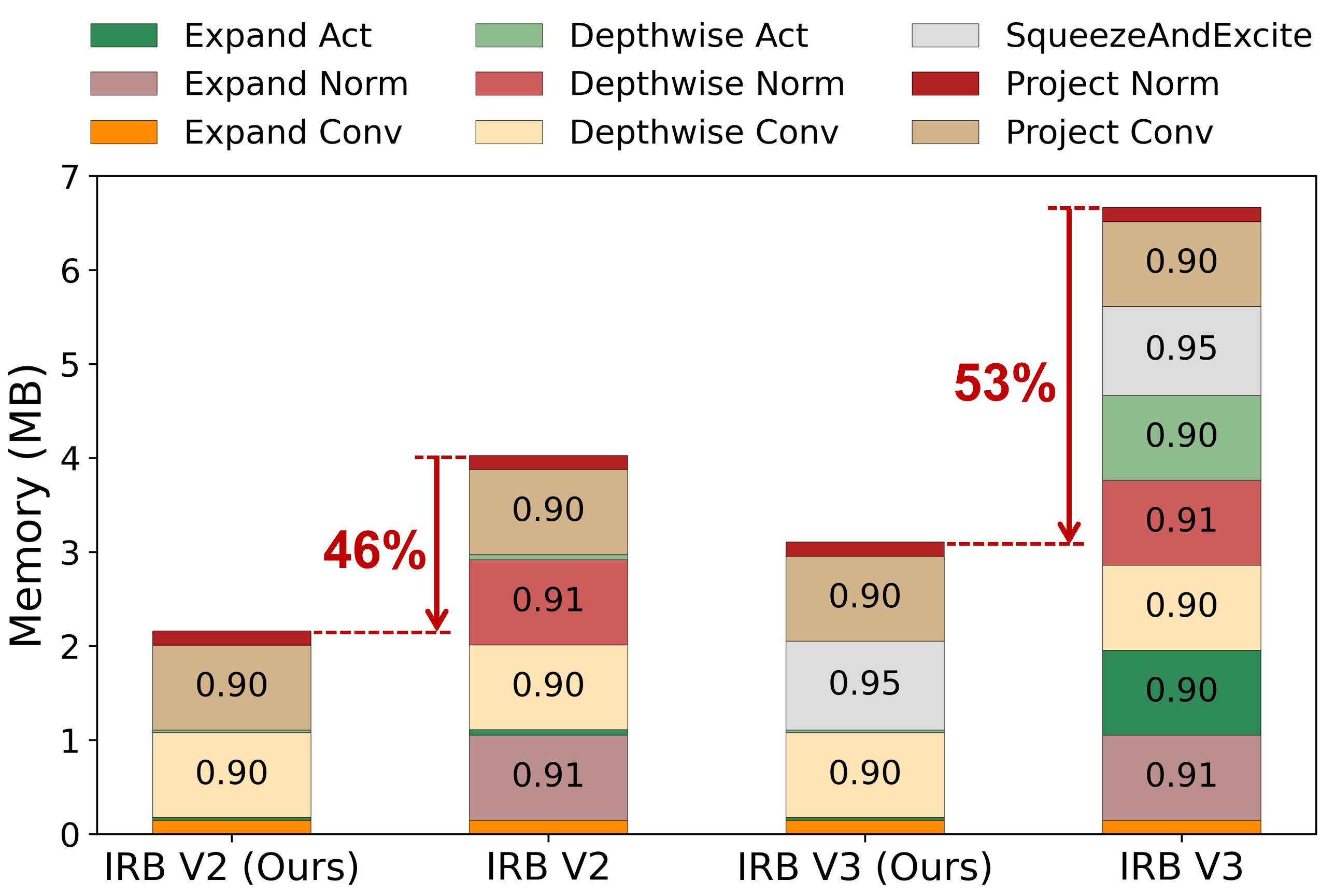}
\caption{
The figure shows stored activations of IRBs for the backward pass.
We set the expansion ratio to 6.
MobileTL reduces memory cost by 46.3\% and 53.3\% for MobileNetV2 and V3 blocks.
}
\label{our-blocks}
\end{figure}

\begin{figure*}[h]
\begin{center}
\centerline{\includegraphics[width=0.9\textwidth]{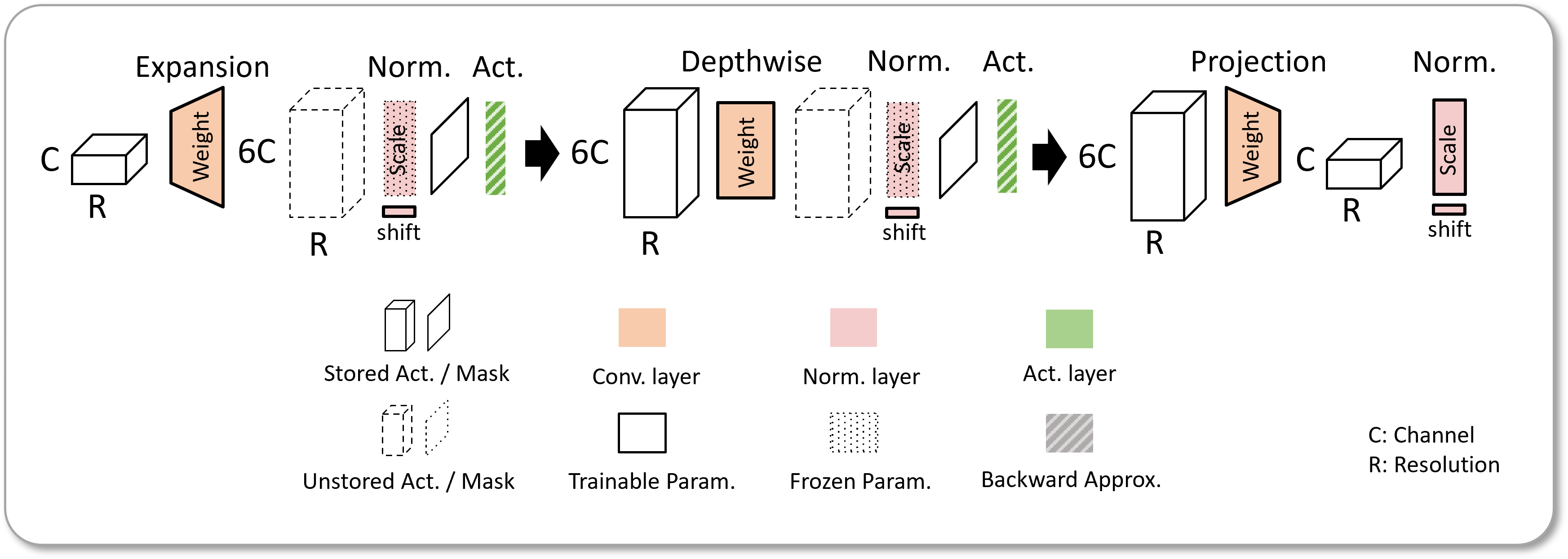}}
\caption{
MobileTL is an efficient training scheme for IRBs.
To avoid storing activation maps for two normalization layers after expansion and depthwise convolution, we only train shifts, and freeze scales and global statistics.
The weights in convolutional layers are trained as usual.
To adapt the distribution to the target dataset, we update the scale, shift, and global statistics for the last normalization layer in the block.
MobileTL approximates the backward function of activation layers, \emph{e.g.}, ReLU6 and Hard-Swish, by a signed function, so only a binary mask is stored for activation backward computing.
Our method reduces the memory consumption by 46.3\% and 53.3\% for MobileNetV2 and V3 IRBs, as shown in Figure \ref{our-blocks}.
}
\label{our-method}
\end{center}
\end{figure*}

On-device transfer learning is challenging due to limited computational resources, often prohibiting fine-tuning models.
Inverted Residual Block (IRB) is one of the prevalent building blocks for models targeting mobile platforms.
An IRB comprises one depthwise convolution, and two pointwise convolutions layers \cite{chollet2017xception} with normalization and activation layers.
IRBs expand the feature maps to higher dimensions using a pointwise convolution (with an expansion ratio of 3, 6, 8 \emph{etc.}), apply a depthwise convolution in the higher dimensional space, and then project the features back into lower dimensions using a pointwise convolution.
Though it is computationally efficient with fewer parameters, an IRB replaces one convolution with a stack of layers resulting in additional memory overhead for storing activation maps during training.
Table \ref{conv-cost} shows that MobileNetV2 and V3 blocks consume $0.913$ and $1.362$ MB in training for storing activation maps, which is $2.98 \times$ and $4.45 \times$ more than for a vanilla convolution block.

To address this issue, we propose a memory-efficient back-propagation method for IRB-based models to enable fine-tuning on edge devices.
As shown in Figure \ref{our-method}, for each intermediary normalization layer, we only update the shift but freeze the scale and global statistics (\textit{i.e.,} mean and variance).
This means we no longer have to store as many activation maps.
The backward pass for memory-intensive activation layers, such as Hard-Swish \cite{howard2019searching} layers, are approximated as signed functions.
The approximation allows us to store a binary mask when propagating the gradient.
To reduce the memory footprint and FLOPs, we compute the gradient and update a few front-end blocks in floating point precision while the rest of the parameters are frozen and quantized during transfer learning.
From our experiments, MobileTL reduces memory usage by 53\% and 46\% for MobileNetV3 and V2 blocks, respectively, and outperforms the baseline under the same memory constraint.


\section{Related Work}
\label{related work}
Low-cost, low-latency, and few-shot deep learning algorithms are key to budget-limited, customized, and data-sensitive use cases.
A large body of research has been proposed to improve training, inference, and transfer learning efficiency. 

\paragraph{Efficient Model and Inference}
Designing an efficient architecture with reduced parameters, memory footprint, and FLOPs has drawn much research attention.
\cite{chollet2017xception} decomposes an over-parameterized and computation-heavy convolution layer into separable convolution layers.
\cite{iandola2016squeezenet, howard2017mobilenets, sandler2018mobilenetv2, zhang2018shufflenet} handcraft efficient building blocks to build models for mobile platforms.
The IRB comprised of depthwise and pointwise convolutions \cite{sandler2018mobilenetv2} is now one of the most prevalent structures for mobile platforms.
\cite{howard2019searching, cai2018proxylessnas} search for model architectures for neural nets from the search space built with efficient building blocks.
To reduce the memory footprint and to boost the latency in inference, pruning \cite{han2015deep} and quantization \cite{zhou2016dorefa, courbariaux2015binaryconnect, dong2019hawq} of model weights are prevalent methods.
Our work addresses efficient \emph{training} on devices and thus is different from the aforementioned work.

\paragraph{Efficient Training}
\cite{wu2018training, zhu2020towards} propose accelerating the training process with low-bitwidth training, thereby having a lower memory footprint.
To save computational cost, \cite{Yang2021moil} skip the forward pass by caching the feature maps and only trains the last layer.
Sparse training techniques \cite{ma2021memory, mostafa2019parameter, dettmers2019sparse} are proposed to update a subset of parameters under a constant resource constraint, but additional overhead for selecting trainable parameters is needed.
\cite{cai2020tinytl, mudrakarta2018k} train lightweight operators and specific parameters, \emph{e.g.}, scales and shifts, to achieve a lower memory footprint and transfer the pre-trained model to the target dataset.
Our method studies the efficient training method for the existing blocks.
We update all parameters in convolution layers in trainable blocks without adding patches or selecting training parameters. 

\paragraph{Transfer Learning}
Our work is closely related to the transfer learning paradigm. 
\cite{mudrakarta2018k} show that training scales and shifts in normalization layers effectively transfers the embedding to the target domain.
\cite{houlsby2019parameter} propose adapter modules to transfer large transformer models to new tasks without global fine-tuning.
\cite{cai2020tinytl} train bias to avoid storing activation maps and add memory-efficient lite-residual modules to recover the accuracy. 
Our work proposes an efficient transfer learning strategy for MobileNet-like models, which is orthogonal to previous work.
In contrast, our method does not alter the model architecture and thus generalizes to any architecture.

\begin{figure}[t]
\centering
\includegraphics[width=52mm]{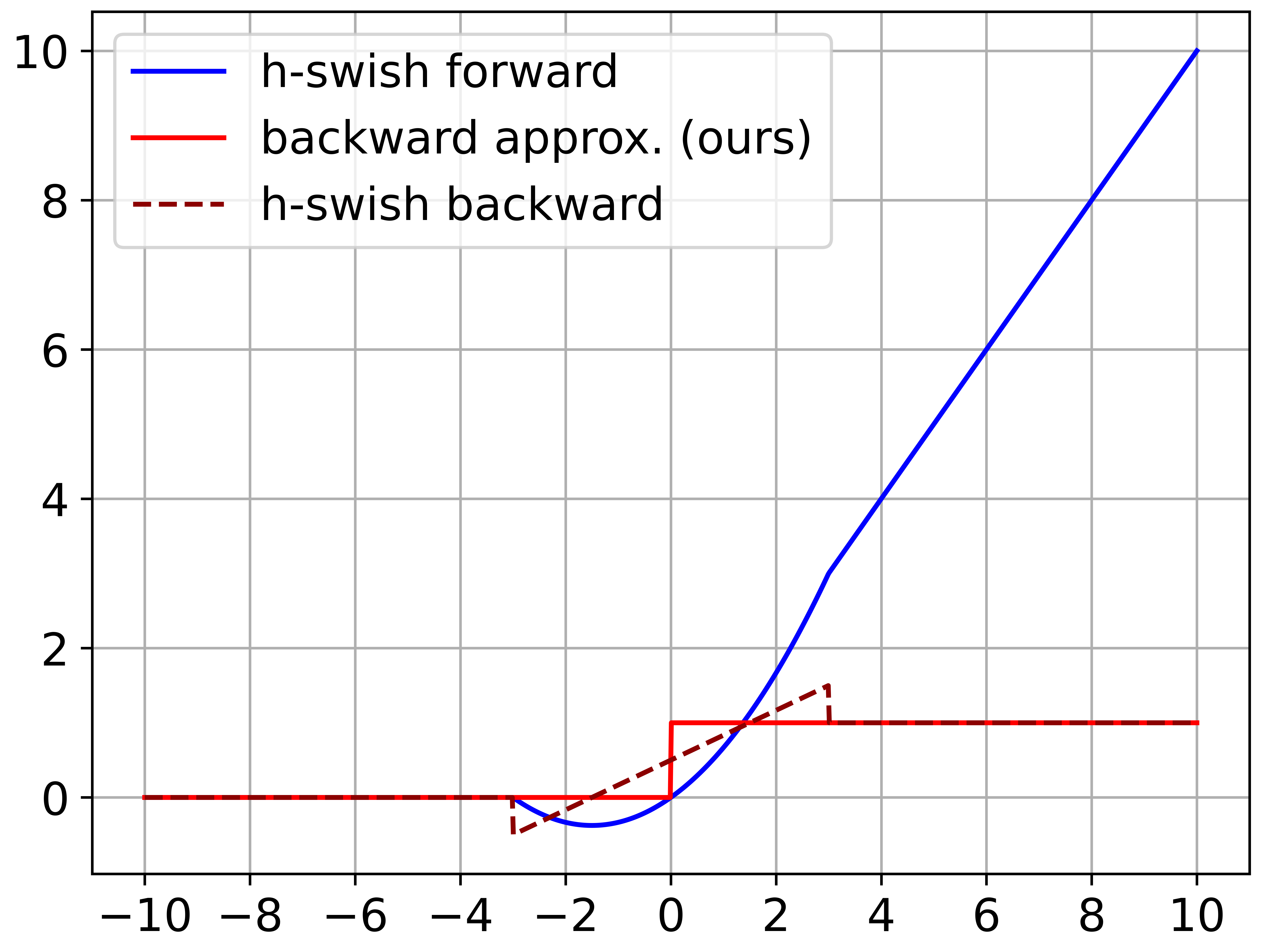}
\caption{
We approximate a Hard-Swish activation with a signed function.
As a result, only a binary mask is involved in the backward computation for a Hard-Swish layer.
}
\label{hswish-bwd}
\end{figure}

\section{MobileTL Overview}
\label{mobiletl}

\subsection{Efficient Transfer Learning}
\begin{figure}[t]
\centering
\includegraphics[width=85mm]{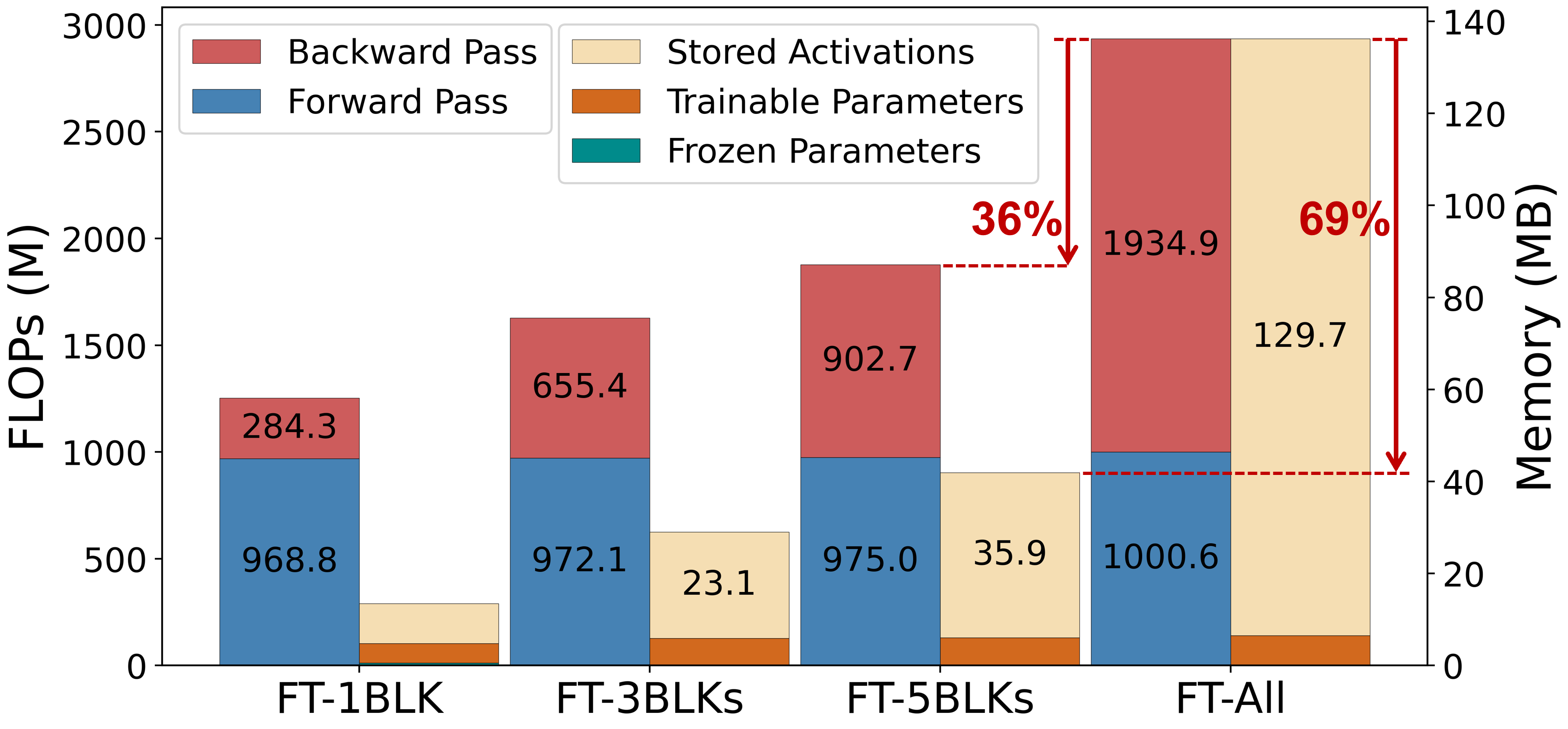}
\caption{
The FLOPs count and memory cost for training MobileNetV3 Small under different settings, \emph{e.g.}, fine-tune 3 blocks (FT-3BLKs), fine-tune all parameters (FT-All). In all settings, the classification layers and feature fusion layers are trained.
}
\label{ft-cost}
\end{figure}

Fine-tuning all parameters can incur a huge computational cost.
Figure \ref{ft-cost} graphs FLOP counts on the left axis and memory on the right axis for fine-tuning MobileNetV3 Small \cite{howard2019searching}.
To perform global fine-tuning, the backward pass takes roughly twice the FLOPs of the forward pass (1934.92 \textit{vs.} 1000.6 MFLOPs), and accumulated activation maps of each layer occupy 129.7 MB (19.9$\times$ model size) based on the chain-rule. 
To avoid global fine-tuning, we decompose a trained model $f(x)$ into $g$ and $h$ functions such that $f(x) = h(g(x))$. 
We assume bottom blocks (close to input) learn primitive features, \emph{e.g.}, corners and edges, which can be shared across different tasks, while top blocks (close to output) recognize entire objects that are more task-specific \cite{zeiler2014visualizing}.
Based on this assumption, we freeze and quantize bottom blocks $g(x)$ on 8-bit precision and only update top blocks $h(x)$ for the target dataset during transfer learning.
Therefore, our method does not propagate the gradient through the entire network and avoids storing activation maps for bottom blocks in $g(x)$.

\begin{table}[t]
\centering
\small
\begin{tabular}{@{}c|c|c|c@{}}
\toprule
Block   & \# Param.      & FLOPs (M)        & Store Act. (MB) \\ \midrule\midrule
Conv    & 230592         & 541.67           & \textbf{0.306 } \\ \midrule
MBV2    & \textbf{21408} & \textbf{51.56}   & 0.913           \\ \midrule
MBV3    & 26136          & 52.91            & 1.362           \\ \bottomrule
\end{tabular}
\caption{
We set the input and output sizes to be $(8, 96, 7, 7)$, and the expansion ratio of IRBs to 1.
The weight filter size for vanilla and depthwise convolution is $5 \times 5$.
All convolution layers are followed by a normalization layer.
IRBs reduce the number of parameters by $10.7 \times$ and FLOPs count by $10.5 \times$ while maintaining the same receptive field, but increase the training memory by $2.9 \times$ and $4.4 \times$, respectively.
}
\label{conv-cost}
\end{table}

\begin{table*}[t]
\centering
\small
\label{tab:activation}
\begin{tabular}{@{}c|c|c|c@{}}
\toprule
Activation & Forward                                 & Backward & Memory   \\ \midrule\midrule
ReLU6    & $\boldsymbol{a}_{i+1}=\text{min}(\text{max}(0, \boldsymbol{a}_i), 6)$                       &  $ \frac{\partial {L}}{\partial{a_{i}}} = \frac{\partial {L}}{\partial{\boldsymbol{a}_{i+1}}} \circ \textbf{1}_{0 \leq \boldsymbol{a}_i \leq 6 }$    & 2 $|\boldsymbol{a}_i|$ \\
Ours   & $\boldsymbol{a}_{i+1}=\text{min}(\text{max}(0, \boldsymbol{a}_i), 6)$                       &  $ \frac{\partial {L}}{\partial{a_{i}}} = \frac{\partial {L}}{\partial{\boldsymbol{a}_{i+1}}} \circ \textbf{1}_{\boldsymbol{a}_i \ge 0}$    &  $|\boldsymbol{a}_i|$ \\ \midrule
Hard-Swish & $\boldsymbol{a}_{i+1}=\boldsymbol{a}_i \circ \frac{\text{ReLU6}(\boldsymbol{a}_i + 3)}{6}$  & $ \frac{\partial {L}}{\partial{a_{i}}} = \frac{\partial {L}}{\partial{\boldsymbol{a}_{i+1}}} \circ (\frac{\text{ReLU6}(\boldsymbol{a}_i+3)}{6} + \boldsymbol{a}_i \circ \frac{\textbf{1}_{-3 \leq \boldsymbol{a}_i \leq 3}}{6}) $    &  32 $|\boldsymbol{a}_i|$ \\ 

Ours & $\boldsymbol{a}_{i+1}=\boldsymbol{a}_i \circ \frac{\text{ReLU6}(\boldsymbol{a}_i + 3)}{6}$  & $ \frac{\partial {L}}{\partial{a_{i}}} = \frac{\partial {L}}{\partial{\boldsymbol{a}_{i+1}}} \circ \textbf{1}_{\boldsymbol{a}_i \ge 0}$    &  $|\boldsymbol{a}_i|$ \\ \bottomrule
\end{tabular}
\caption{We approximate the Hard-Swish and ReLU6 backward pass by a signed function. Instead of storing the full activation map, we store a compact binary mask for the backward pass, thereby reducing intermediary memory cost. $|\cdot|$ denotes the number of elements. The memory is calculated in bits.}
\end{table*}

\subsection{Efficient Transfer Learning Block}
\label{transfer_learning_block}
\emph{We argue that updating intermediary normalization layers in the IRB is ineffective in the transfer learning paradigm} because it consumes a large amount of memory but without producing significant accuracy improvements.
As a result, for transfer learning, MobileTL proposes to simplify the IRB by avoiding storing activation maps for normalization layers in between.
More specifically, given a batch normalization (BN) layer in evaluation mode, \emph{i.e.}, $\boldsymbol{\mu_{\boldsymbol{x}}}_i$ and $\boldsymbol{\sigma_{\boldsymbol{x}}}_i$ are frozen.
\begin{equation}
\begin{aligned}
& \text{BN}  (\boldsymbol{x}_i, \boldsymbol{\gamma}_i, \boldsymbol{\beta}_i, \boldsymbol{\mu_{\boldsymbol{x}}}_i, \boldsymbol{\sigma_{\boldsymbol{x}}}_i)
= \boldsymbol{\gamma}_i
\frac{\boldsymbol{x}_i - \boldsymbol{\mu}_i}{\boldsymbol{\sigma}_{\boldsymbol{x}_i}} + \boldsymbol{\beta}_i
= \Bar{\boldsymbol{x}}_i
\text{.}
\end{aligned}
\end{equation}

The gradient with respect to the scale $\boldsymbol{\gamma}_{i}$  and the shift $\boldsymbol{\beta}_{i}$ are
\begin{equation}
\label{bn_bwd}
\begin{aligned}
\frac{\partial L}{\partial \boldsymbol{\gamma}_{i}} = \frac{\partial L}{\partial \Bar{\boldsymbol{x}}_{i}} \frac{\boldsymbol{x}_i - \boldsymbol{\mu}_i}{\boldsymbol{\sigma}_{\boldsymbol{x}_i}}
, \;
\frac{\partial L}{\partial \boldsymbol{\beta}_{i}} = 
\frac{\partial L}{\partial \Bar{\boldsymbol{x}}_{i}}
\text{.}
\end{aligned}
\end{equation}

From Eq. \ref{bn_bwd}, we have to store the activation $\boldsymbol{x}_{i}$ to compute the gradient for $\boldsymbol{\gamma}_{i}$.
Therefore, to avoid accumulating activation maps in memory, we freeze scales and global statistics for intermediary normalization layers when training IRBs, as Fig. \ref{our-method} shows.
Both normalization layers normalize inputs with pre-trained statistics.
To recover the distribution difference between the pre-training dataset and target datasets, we update shifts in both intermediary normalization layers.
We keep the global mean and variance in the final normalization layer updating during training, and both its scale and shift are trained for adapting to the distribution.

\label{active_approx}
Though Hard-Swish \cite{howard2019searching} can boost the performance, it requires storing activation maps in memory during the backward pass, as shown in Table \ref{tab:activation}.
We propose approximating its backward computation as a signed function.
As a result, only a binary mask is stored in memory for later backward computation.
Figure \ref{hswish-bwd} depicts the forward and backward mapping of our implementation for Hard-Swish.

Training $L$ layers in a neural network with backward-approximated Hard-swish activation functions, we derive a theoretical bound with training steps ($T$) and the number of approximated layers ($L$).
We use standard stochastic gradient descent with a learning rate $\lambda$ and assume the magnitude of the gradient is bounded by $G$, and the number of elements in the output from each layer is bounded by $N$.
We also assume that the network function $F(\boldsymbol{\cdot})$ is Lipschitz continuous, \emph{i.e.,} $\forall \boldsymbol{x}, \boldsymbol{y} \in \text{dom} F $, $ \|F(\boldsymbol{x}) - F(\boldsymbol{y})\|_2 \leq M\|\boldsymbol{x}-\boldsymbol{y}\|_2$,
We derive an error bound for the loss between the weights trained with MobileTL and the original weights.
We let the weights in trainable $L$ layers from MobileTL be $\boldsymbol{\Tilde{W}} = (\boldsymbol{\Tilde{w}}_1, \boldsymbol{\Tilde{w}}_2, ..., \boldsymbol{\Tilde{w}}_L)$, and the original weights are $\boldsymbol{W} = (\boldsymbol{w}_1, \boldsymbol{w}_2, ..., \boldsymbol{w}_L)$.
\begin{theorem}
\label{th}
Given trainable $L$ layers in a neural network with Hard-swish activation functions, whose backward calculation is approximated with a signed function. If we train the $L$ layers for $T$ steps, then the loss distance between $\|F(\boldsymbol{\Tilde{W}}^T) - F(\boldsymbol{W}^T)\|$ is bounded by $\lambda M T G \left(\frac{\Psi (1-\Psi^L)}{1-\Psi} + \frac{\Tilde{\Psi} (1-\Tilde{\Psi}^L)}{1-\Tilde{\Psi}}\right)$, where $\Psi = \frac{3}{2} \sqrt{N}G, \Tilde{\Psi} = \sqrt{N}G$, and $M$ is the constant from the Lipschitz continuous property of $F(\boldsymbol{\cdot})$.
\end{theorem}
The proof of Theorem \ref{th} can be found in the Supplementary Material.
Theorem \ref{th} shows that MobileTL transfers the pre-train weights to target datasets without incurring accuracy drop since on-device datasets are orders of magnitude smaller than pre-trained datasets (necessary updates $T$ is
small) and we approximate the Hard-swish layers only in trainable blocks ($L$ is small).
Figure \ref{ft-cost} shows FT-3BLKs with MobileTL close to the knee point of the curves.

\section{Experimental Results}

\subsection{Experiment Setup}

\paragraph{Model Profiling}
We develop an analytical profiling tool to theoretically calculate the number of floating-point operations and memory footprint for both forward and backward passes.
Floating-point operations can generalize to more operations such as normalization and activation layers and are not limited to multiplication–accumulate operations (\emph{i.e.}, MACs).
For simplicity, we approximate all operations as one FLOP, although not all floating-point operators consume the same energy and number of clock cycles. For example, exponentiation and square root require many more cycles than addition. However, we still approximate them as a single FLOP as done by most analytical tools.
To calculate the memory footprint, we consider all training components, including trainable and frozen parameters, accumulated activation maps, temporary memory for matrix multiplication, and residual connections.
By considering all intermediary variables and operations, our profiling tool can better represent a complete training scheme, whereas many techniques only consider the model size.

\begin{table}[t]
\centering
\small
\begin{tabular}{@{}cccc@{}}
\toprule
Methods                                   &  Mem.            &  FLOPs            &  Train  \\ 
                                          &  (MB)            &  (M)              &  Param. \\   \midrule\midrule
FT-All                                    & 382.7            & 16,205.0          & 2,927,612          \\   \midrule
FT-BN                                     & 189.9            & 11,066.1          & 162,596            \\
FT-Bias                                   & 30.6             & 10,446.5          & 145,348            \\ 
FT-3BLKs                                  & 40.5             & 7,728.0           & 1,695,972          \\ 
FT-Last                                   & 29.0             & 5,325.9           & 128,100            \\ \midrule
TinyTL-B                                  & 31               & 10,446.5          & 145,348            \\ 
TinyTL-L\textsuperscript{\textdagger}     & 32               & 13,505.9          & 1,944,516          \\ 
TinyTL-L-B\textsuperscript{\textdagger}   & 37               & 14,087.6          & 1,959,748          \\ \midrule
\textbf{MobileTL-3BLKs\textsuperscript{*}}         & \textbf{33.7}    & \textbf{7,699.0}  & \textbf{1,691,364} \\ \bottomrule
\end{tabular}

\footnotesize{\textsuperscript{\textdagger} uses model patches during fine-tuning.\textsuperscript{*} denotes Pareto-optimal }
\caption{
We investigate different finetuning strategies with Proxyless Mobile on CIFAR100.
All methods predicated with ``FT" represent vanilla fine-tuning on the corresponding layer type.
For example, FT-Bias represents fine-tuning only the bias term of each linear.
MobileTL is Pareto-optimal among all methods across various datasets (\textit{c.f.} Figure \ref{main-results}).
}
\label{comp-tinytl}
\end{table}

\paragraph{Dataset}
We apply transfer learning on multiple image classification tasks. Similar to prior work \cite{kornblith2019better, houlsby2019parameter, cai2020tinytl},
we begin with an ImageNet \cite{deng2009imagenet} pre-trained model, and transfer to eight downstream image classification datasets, including Cars \cite{krause20133d},
Flowers \cite{nilsback2008automated}, Aircraft \cite{maji2013fine}, CUB-200 \cite{wah2011caltech}, Pets \cite{parkhi2012cats}, Food \cite{bossard2014food}, CIFAR10 \cite{krizhevsky2009learning}, and CIFAR100 \cite{krizhevsky2009learning}.

\paragraph{Model Architecture}
We apply MobileTL to a variety of IRB-based models, including MobileNetV2 \cite{sandler2018mobilenetv2}, MobileNetV3\cite{howard2019searching}, and Proxyless Mobile \cite{cai2018proxylessnas}.
Though MobileTL targets models built with IRBs, we illustrate MobileTL's flexibility in the analysis section, by extending our method to models built with conventional convolution blocks such as ResNet18 and ResNet50 \cite{he2016deep}.

\begin{figure*}[t]
\centering
\centerline{\includegraphics[width=\textwidth]{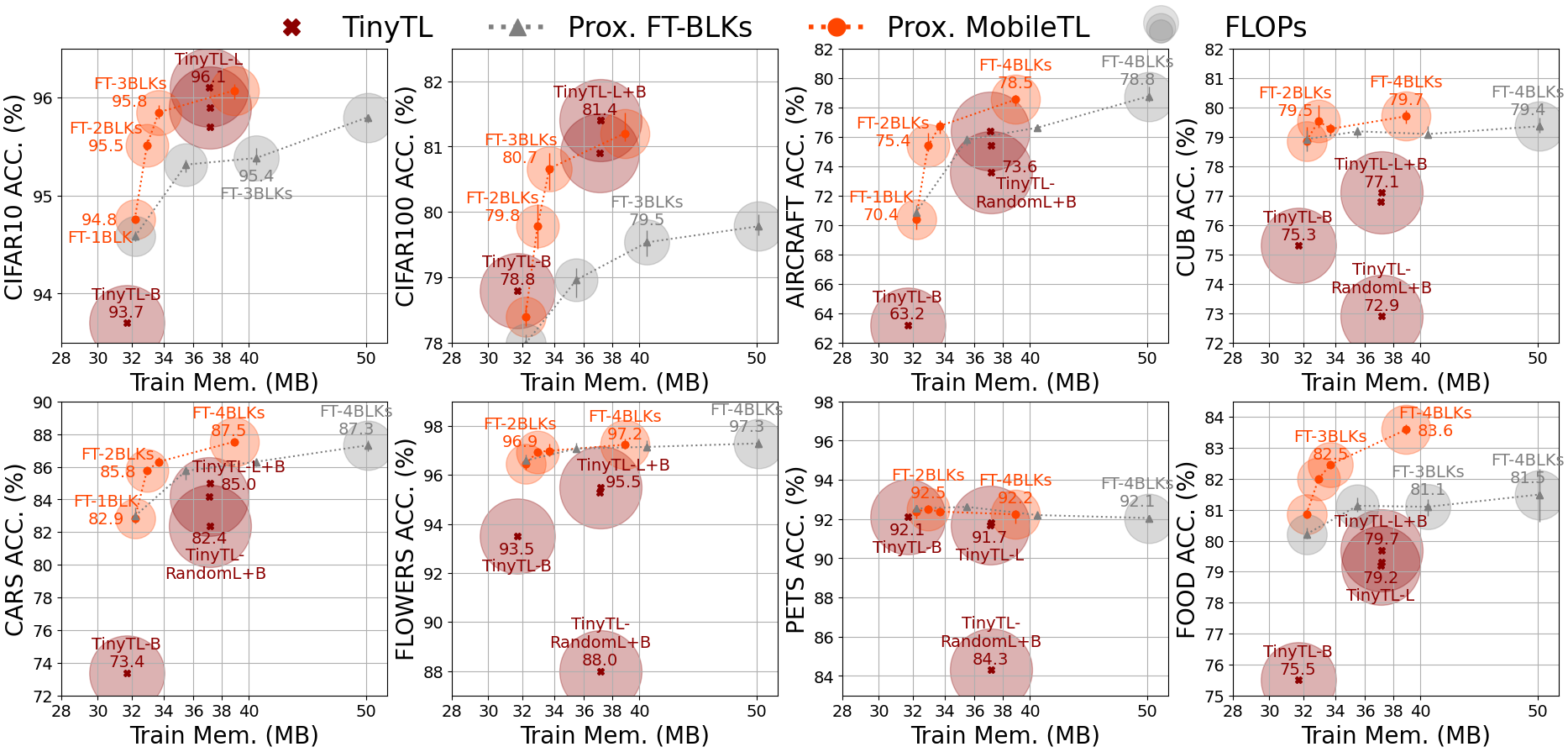}}
\caption{We experiment MobileTL with Proxyless Mobile \cite{cai2018proxylessnas} and transfer to eight downstream tasks. We compare our method (Prox. MobileTL in orange) with vanilla fine-tuning a few IRBs of the model (Prox. FT-BLKs in grey) and TinyTL \cite{cai2020tinytl}. MobileTL maintains Pareto optimality for all datasets and improves accuracy over TinyTL in six out of eight datasets.}
\label{main-results}
\end{figure*}

\paragraph{Training Details}
For fair comparison, we follow the hyper-parameters and settings in TinyTL \cite{cai2020tinytl} where they train for 50 epochs with a batch size of 8 on a single GPU.
We use the Adam optimizer and cosine annealing for all experiments, however, the initial learning rate is slightly tuned for each dataset and model.
The classification layers are trained in all settings, and fusion layers are trained in block-wise fine-tuning.
We ran our experiment using four random seeds, and average the results.

\begin{figure*}[t]
\centering
\includegraphics[width=\textwidth]{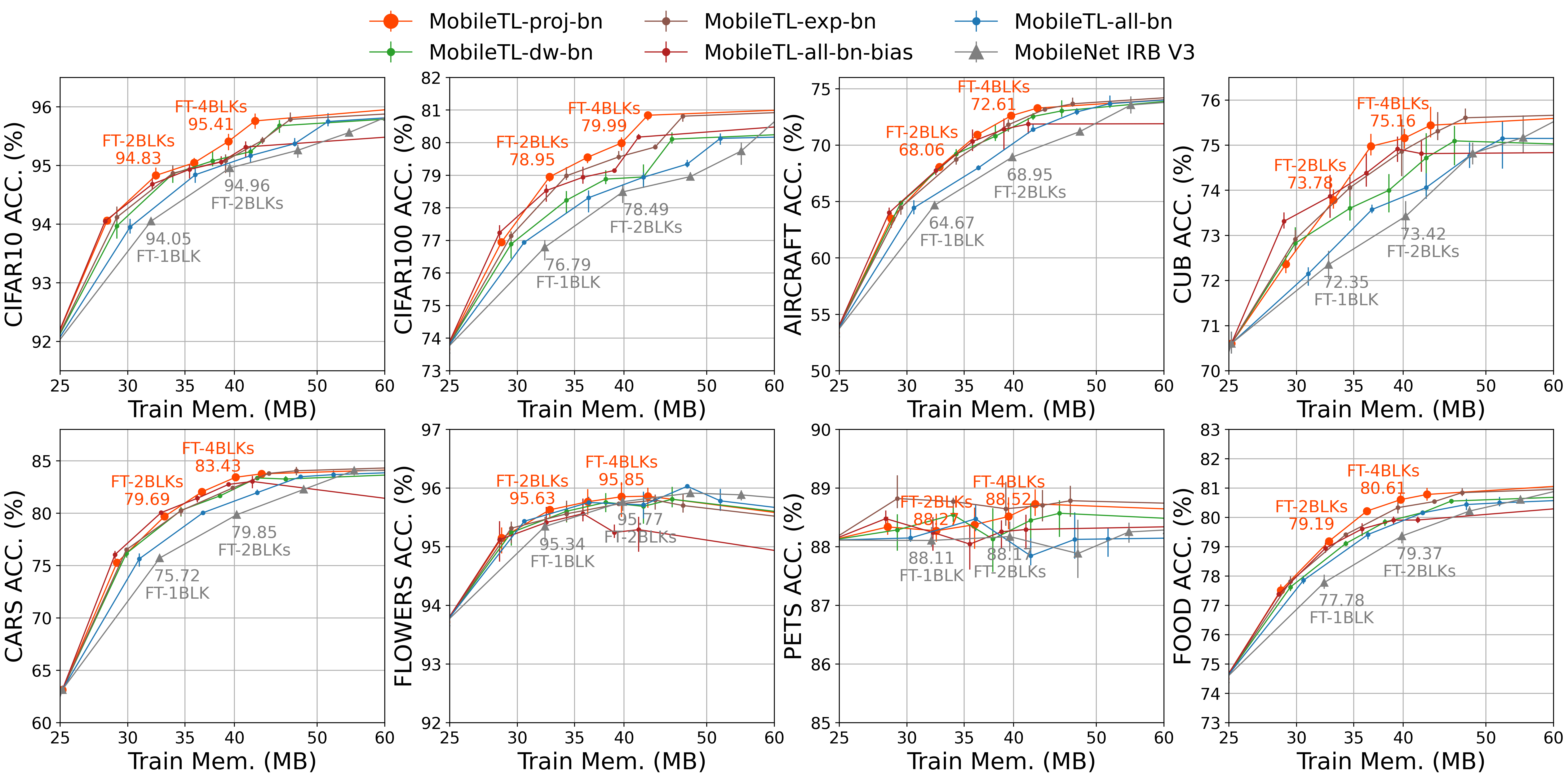}
\caption{
MobileTL-proj-bn is Pareto-optimal for MobileNetV3 Small across different settings.
The last normalization layer following behind the projection layer takes less memory for training while adapting global statics to the target dataset.
MobileTL-all-bn-shift (red) has the least memory consumption but with degraded accuracy.
The results show that training shifts only in normalization layers fails to adapt the weights to the target domain.
}
\label{ablation}
\end{figure*}

\subsection{Efficient Transfer Learning with MobileTL}
Table \ref{comp-tinytl} presents different fine-tuning strategies on Proxyless Mobile \cite{cai2018proxylessnas}.
TinyTL-B \cite{cai2020tinytl} only trains bias and avoids accumulating activation maps in memory.
However, the gradient propagates to biases in the whole network, thereby requiring more FLOPs.
In order to recover the accuracy, the lightweight patch in residual connection is proposed to train with the model (TinyTL-L-B).
However, these patches introduce new parameters to the model.
To best recover the performance for transfer learning, these patches need to be trained with the model on the large-scale pre-task dataset.
We adopt a different approach by starting with fine-tuning a few IRBs, \emph{e.g.}, 3 blocks of the model, and then applying MobileTL to trainable IRBs.
MobileTL reduces memory usage by $16.7\%$ when compared to vanilla fine-tuning of three IRBs and reduces FLOPs by $52 \%$ when compared to global fine-tuning.

Figure \ref{main-results} depicts the accuracy versus memory footprint for transferring an ImageNet pre-trained Proxyless Mobile to eight downstream tasks.
The radius of a circle represents the number of FLOPs, and therefore a smaller area means a smaller FLOP count.
Compared with the baselines, MobileTL is Pareto-optimal under the same memory constraint for widely adopted datasets such as CIFAR10 \cite{krizhevsky2009learning}, Aircraft \cite{maji2013fine}, CUB-200 \cite{wah2011caltech}, \textit{etc}.
For CIFAR100 \cite{krizhevsky2009learning}, MobileTL has comparable performance to TinyTL \cite{cai2020tinytl} but with lower FLOPs as well as lower latency on edge devices (\textit{c.f.} Table \ref{ablation-device}).
In our experiments, MobileTL outperforms the vanilla version in CIFAR10 and CIFAR100, illustrating that our method transfers the pre-trained model to the target dataset with lower memory costs.
Our experiments also show that the pre-trained normalization statistics from the large pre-task dataset benefit downstream tasks.
We can leverage this by only training the shift parameter and maintaining the original normalization statistics.
MobileTL outperforms vanilla fine-tuning by $0.47 \%$ and $1.13\%$ in accuracy and has lower memory costs when fine-tuning three blocks in CIFAR10 and CIFAR100 respectively.
%

\section{Analysis}

\subsection{Normalization Layers in IRBs}
\label{section-norm-layers}
We study the effect of training different normalization layers for MobileTL and show the results in Figure \ref{ablation}.
In this experiment, we adopt MobileNetV3 Small \cite{howard2019searching}.
For this ablation, the batch normalization layer following each expansion, depthwise, and pointwise convolution is trained in isolation (denoted in Figure \ref{ablation} by -exp-bn, -dw-bn, and -proj-bn, respectively).
For example, MobileTL-proj-bn fully updates the last normalization layer after the projection layer, while the other two normalization layers only update shifts.
%
MobileTL-all-bn-shift only updates shifts and freezes scales and global statistics for all normalization layers in trainable IRBs.
In contrast, MobileTL-all-bn fully updates scales, shifts, and global statistics for all normalization layers.
In all MobileTL settings, Hard-Swish layers in trainable IRBs are approximated as a signed function in the backward pass.

In our experiments, we show that MobileTL-proj-bn (orange) is the most memory-efficient and Pareto-optimal.
Training the last normalization layer involves less memory while adapting the pre-trained weights to the target dataset.
MobileTL-all-bn-shift (red) has the least memory consumption but with degraded accuracy, which shows that \emph{training shifts only in normalization layers fails to adapt the weights to the target domain.}

\subsection{Training with Efficient Operators}
\label{section-efficient-ops}
A natural way to reduce training memory is by substituting Hard-Swish with ReLU or removing Squeeze-and-Excitation layers whose backward update is very memory intensive (\emph{c.f.} Fig. \ref{our-blocks}).
However, pre-trained weights depend on the model architectures and are sensitive to operators in the network.
In Table \ref{ablation-relu}, we show that MobileTL is superior to these naive techniques.
Furthermore, MobileTL does not alter the network structure and therefore avoids the performance drop when transferring pre-trained weights to the target dataset.

\begin{table}[t]
    \centering
    \small
    \begin{tabular}{@{}c|c|c@{}}
    \toprule
    IRB V3     & Mem. (MB)      &  CIFAR10 (\%)  \\ \midrule \midrule
    Vanilla    & 47.4          &  95.2         \\ \midrule
    remove-SE      & 43.6          &  94.3         \\
    ReLU       & 41.7          &  94.6         \\ 
    MobileTL   & \textbf{35.8} & \textbf{95.0} \\
    \bottomrule
    \end{tabular}
    \caption{We remove SE Layers (the second row) or replace H-Swish activation function with ReLU (the third row) for IRBs in MobileNetV3 Small. Although they reduce memory footprint, they lead to lower accuracy when transferring ImageNet to CIFAR10. MobileTL's accuracy in the last row approaches the vanilla fine-tuning in the first row. 
    }
    \label{ablation-relu}
\end{table}

\subsection{Model Patches for Transfer Learning}
\label{section-patches}
MobileTL is orthogonal to previous work \cite{houlsby2019parameter, cai2020tinytl} that adds lightweight patches to the model and transfers the patches to the target dataset.
In Table \ref{ablation-patch}, we transfer Proxyless Mobile from ImageNet-pre-trained weights to CIFAR10.
We fine-tune the three top blocks (close to output) of the model.
The first row is the vanilla block-wise fine-tuning approach.
The following three rows correspond to MobileTL.
We add model patches in the residual connections for the last two rows in the table.
We adopt the lite-residual module as our experimental patches proposed in \cite{cai2020tinytl} with resolution down-sampling, group normalization layers \cite{wu2018group}, and group convolutions.
The patches are without pre-trained weights and are randomly initialized. 
The results show that patches present \emph{additional $39.9\%$ trainable parameters} while having marginal improvement training with main blocks.
In contrast, MobileTL \emph{reduces the memory footprint while keeping the accuracy without the need of adding additional modules or increasing parameters to the model}.

\begin{table}[t]
    \small
    \centering
    \begin{tabular}{@{}c|cc|c|c|c@{}}
    \toprule
    Mobile     & Main       & Res.        &  Train  & Mem. &  CIFAR10 \\ 
       TL      & Blk        & Patch       &  Param. & (MB) &  (\%)    \\ \midrule \midrule
               & \checkmark &             & 1,580,682 & 40.1 &  95.4    \\ \midrule
    \checkmark & \checkmark &             & \textbf{1,576,074} & \textbf{33.2} &  \textbf{95.8}    \\ 
    \checkmark & \checkmark & \checkmark  & 2,211,466 & 35.8 &  95.8    \\
    \checkmark &   frozen   & \checkmark  & 1,060,362 & 32.3 &  94.4    \\ 

    \bottomrule
    \end{tabular}
    \caption{We study the effectiveness of model patches. We transfer Proxyless Mobile from ImageNet to CIFAR10. The lightweight patches in the residual connection bring marginal improvement when training with main blocks. In contrast, MobileTL reduces the memory cost without incurring accuracy drop.}
    \label{ablation-patch}
\end{table}

\begin{table}[t]
    \small
    \centering
    \begin{tabular}{@{}c|cc@{}}
    \toprule
    Device                & \multicolumn{1}{c|}{Method}     & Latency (s) \\ \toprule
    \multirow{4}{*}{Nano} & \multicolumn{1}{c|}{FT-All}     & 0.235  \\
                          & \multicolumn{1}{c|}{FT-BN}      & 0.138  \\
                          & \multicolumn{1}{c|}{FT-Bias}    & 0.130  \\
                          & \multicolumn{1}{c|}{\textbf{MobileTL-3BLKs}} & \textbf{0.114}  \\ \midrule
    \multirow{3}{*}{RPI4} & \multicolumn{1}{c|}{FT-All}     & 2.465   \\
                          & \multicolumn{1}{c|}{FT-BN}      & 1.894   \\
                          & \multicolumn{1}{c|}{FT-Bias}    & 1.818   \\
                          & \multicolumn{1}{c|}{\textbf{MobileTL-3BLKs}} & \textbf{1.344}   \\ \bottomrule
    \end{tabular}
    \caption{We deploy MobileTL on MobileNetV3 Small and measure the average latency for a training step (forward and backward pass) on a NVIDIA JETSON NANO, and a Raspberry PI 4. The latency is measured in seconds. Batch size is set to 8 and input size is $224 \times 224$. The experimental models do not have patches in all settings.}
    \label{ablation-device}
\end{table}

\begin{figure}[h!]
\centering
\includegraphics[width=80mm]{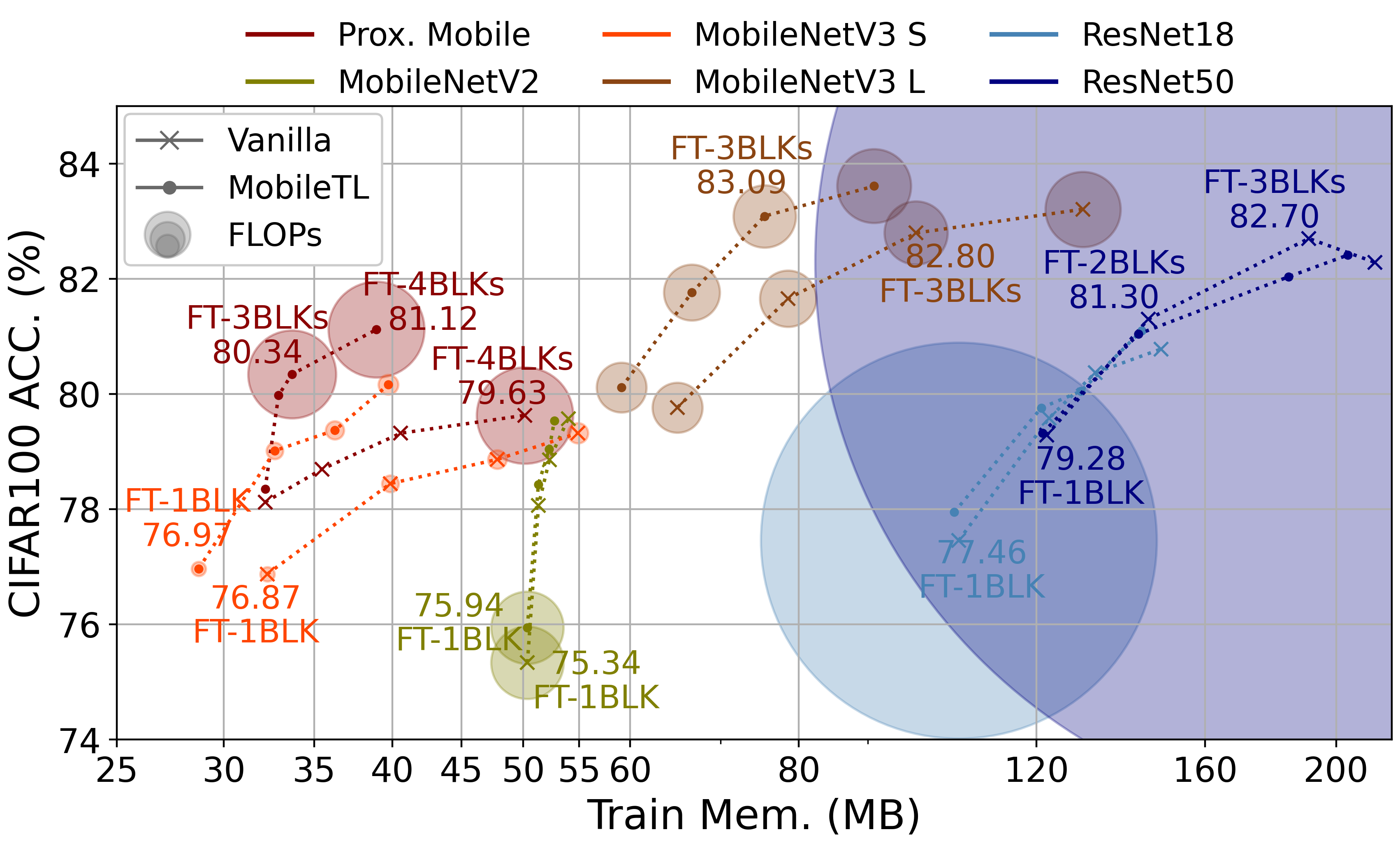}
\caption{We generalize MobileTL to different architectures that built with IRBs, and models built with convolution blocks such as ResNet. The radius corresponds to FLOP count for fine-tuning. MobileTL pushes several models to the Pareto front.}
\label{ablation-models}
\end{figure}

\subsection{Deployment on Edge Devices}
\label{section-deploy}
To demonstrate practical feasibility, we deploy our method to edge devices.
We experiment with Raspberry PI4 model B with Quad core ARM Cortex-A72 64-bit and 4 GB RAM, and NVIDIA JETSON NANO with 128-core GPU, Quad-core ARM Cortex-A57, and 2 GB RAM.
We run our models with PyTorch framework on two devices, while on Raspberry PI is the CPU-only version.
Batch size is set to 8 and input resolution is $224 \times 224$ with output 10 classes. 
We measure the latency of forward and backward passes with random data.
%
The latency is the average of 1000 training steps and is reported in seconds.
The model is pre-loaded to the memory and run for a few warm-up steps before measuring.
We report the result in Table \ref{ablation-device}, which shows that MobileTL reduces the latency by $2.06 \times$ on Nano and $1.82 \times$ on Raspberry PI when compared with global fine-tuning.

\subsection{Generalizing to Other Network Architectures}
\label{section-architecture}
MobileTL does not alter the model architecture and therefore it generalizes to various models.
As shown in Figure \ref{ablation-models}, we apply MobileTL to different architectures based on IRBs, such as MobileNetV2 \cite{sandler2018mobilenetv2}, MobileNetV3 Small and Large \cite{howard2019searching}, Proxyless Mobile \cite{cai2018proxylessnas}.
Additionally, we extend MobileTL to models built with convolution blocks such as ResNet18 and ResNet50 \cite{he2016deep} for comparison.
Figure \ref{ablation-models} depicts memory footprint versus accuracy of different models on CIFAR10 dataset. 
The radius corresponds to the FLOP count for fine-tuning.
MobileTL pushes several models to the knee point, and Proxyless Mobile with MobileTL is Pareto-optimal.
%
%

\section{Conclusion}
We present MobileTL, a new on-device transfer learning method that is memory- and computation-efficient.
MobileTL reduces the memory footprint for separable convolution blocks by freezing the intermediary normalization layers and approximating activation layers in blocks during the backward pass.
To reduce the FLOP counts, we only propagate the gradient through a few trainable top blocks (close to output) in the model to enable fine-tuning on small edge devices.
We show the proposed method generalizes different architectures without re-training weights on the large pre-task dataset since it does not require adding patches to the model or altering the architecture.
Our intensive ablation studies demonstrate the effectiveness and efficiency of MobileTL.

\section*{Acknowledgements}
This research was supported in part by NSF CCF Grant No. 2107085, NSF CSR Grant No. 1815780, and the UT Cockrell School of Engineering Doctoral Fellowship. Additionally, we thank Po-han Li for his help with the proof.

\bibliography{aaai23}

\appendix
\clearpage

\section{Supplementary Material}

We give a theoretical analysis of the output bound of MobileTL in the supplementary material.

\subsection{Proposition}
Let us bound the backward approximation error for a weight matrix in the $i$th layer followed by a Hard-swish activation function.
\begin{equation}
\begin{aligned}
 \boldsymbol{x}_i \boldsymbol{w}_i =\boldsymbol{a}_i
\text{.}
\end{aligned}
\end{equation}
The gradient with respect to the weight matrix $\boldsymbol{w}_i$  and the input matrix $\boldsymbol{x}_i$ are

\begin{equation}
\begin{aligned}
\frac{\partial \boldsymbol{a}_{i}}{\partial \boldsymbol{w}_{i}} =  \boldsymbol{x}_i
, \;
\frac{\partial \boldsymbol{a}_{i}}{\partial \boldsymbol{x}_{i}} =  {\boldsymbol{w}_i}^{\top}
\text{.}
\end{aligned}
\end{equation}
The outputs from the linear layer are passed through a Hard-swish activation function.
\begin{equation}
\text{H-swish}(\boldsymbol{a}_i) = h(\boldsymbol{a}_i) =\boldsymbol{a}_i \circ \frac{\text{ReLU6}(\boldsymbol{a}_i + 3)}{6} = \boldsymbol{x}_{i+1}
\text{.}
\end{equation}
The backward computation of the Hard-swish function is
\begin{equation}
h'(\boldsymbol{a}_i) = \frac{\partial {\boldsymbol{x}_{i+1}}}{\partial{\boldsymbol{a}_{i}}} = (\frac{\text{ReLU6}(\boldsymbol{a}_i+3)}{6} + \boldsymbol{a}_i \circ \frac{\boldsymbol{1}_{-3 \leq \boldsymbol{a}_i \leq 3}}{6})
\text{.}
\end{equation}
Then, we can derive the chain rule to calculate the gradient with respect to the weight matrix $\boldsymbol{w}_i$ and the input matrix $\boldsymbol{x}_i$ as
\begin{equation}
\frac{\partial L}{\partial \boldsymbol{w}_i} = \frac{\partial L}{\partial \boldsymbol{x}_{i+1}} \frac{\partial \boldsymbol{x}_{i+1}}{\partial \boldsymbol{a}_{i}} \frac{\partial \boldsymbol{a}_{i}}{\partial \boldsymbol{w}_{i}}
= \frac{\partial L}{\partial \boldsymbol{x}_{i+1}} h'(\boldsymbol{a}_i) \frac{\partial \boldsymbol{a}_{i}}{\partial \boldsymbol{w}_{i}}, 
\end{equation}
and
\begin{equation}
\frac{\partial L}{\partial \boldsymbol{x}_i} = \frac{\partial L}{\partial \boldsymbol{x}_{i+1}} \frac{\partial \boldsymbol{x}_{i+1}}{\partial \boldsymbol{a}_{i}} \frac{\partial \boldsymbol{a}_{i}}{\partial \boldsymbol{x}_{i}}
= \frac{\partial L}{\partial \boldsymbol{x}_{i+1}} h'(\boldsymbol{a}_i) \frac{\partial \boldsymbol{a}_{i}}{\partial \boldsymbol{x}_{i}}
\text{.}
\end{equation}
We approximate the backward computation of the hardswish activation function as a signed function
\begin{equation}
\Tilde{h}'(\boldsymbol{a}_i) = \text{Approx.}(\frac{\partial {\boldsymbol{x}_{i+1}}}{\partial{\boldsymbol{a}_{i}}}) = \boldsymbol{1}_{\boldsymbol{a}_i \ge 0}
\text{.}
\end{equation}

\subsection{Proof of the Theorem 1}
Suppose we have a $L$-layers neural net, \emph{i.e.}, $\boldsymbol{x}_L$ is the output of the network, and we use standard stochastic gradient descent with a learning rate $\lambda$ to update our weight $\boldsymbol{w}_i, i < L$ for the $i$th layer in training, such that
\begin{equation}
\begin{aligned}
    \boldsymbol{w}^{t+1}_i & = \boldsymbol{w}^t_i 
    \\
    & - \lambda  \frac{\partial L}{\partial \boldsymbol{x}_{L}} h'(\boldsymbol{a}_{L-1}) \frac{\partial \boldsymbol{a}_{L-1}}{\partial \boldsymbol{x}_{L-1}} h'(\boldsymbol{a}_{L-2}) ... h'(\boldsymbol{a}_i) \frac{\partial \boldsymbol{a}_{i}}{\partial \boldsymbol{w}_{i}} 
\text{.}
\end{aligned}
\end{equation}
For MobileTL, we replace the following $h'(\boldsymbol{a}_i)$ with $\Tilde{h}'(\boldsymbol{a}_i)$ in training 
\begin{equation}
\begin{aligned}
    \boldsymbol{\Tilde{w}}^{t+1}_i & = \boldsymbol{w}^t_1 
    \\
    & - \lambda  \frac{\partial L}{\partial \boldsymbol{x}_{L}} \Tilde{h}'(\boldsymbol{a}_{L-1}) \frac{\partial \boldsymbol{a}_{L-1}}{\partial \boldsymbol{x}_{L-1}}\Tilde{h}'(\boldsymbol{a}_{L-2}) ... \Tilde{h}'(\boldsymbol{a}_i)\frac{\partial \boldsymbol{a}_{i}}{\partial \boldsymbol{w}_{i}}
\text{.}
\end{aligned}
\end{equation}

We assume that the magnitude of the gradient is bounded by $G$, and the number of elements in the output from each layer is bounded by $N$.
$\|\boldsymbol{x}\|$ denotes the Frobenius norm of $\boldsymbol{x}$.
According to Fig. \ref{hswish-bwd}, for any point $x$ on the x-axis, the output from the functions, \emph{i.e.}, $|h(x)|$ and $|\Tilde{h}(x)|$ are bounded by $\frac{3}{2}$ and $1$, respectively.

\begin{proposition}
\label{h_dist}
Suppose $h'(\boldsymbol{a}_{i}) \in R^{{n_1}\times{n_2}}$ and $\frac{\partial \boldsymbol{a}_{i}}{\partial \boldsymbol{x}_{i}} \in R^{{n_2}\times{n_3}}$, then $\| h'(\boldsymbol{a}_{i}) \frac{\partial \boldsymbol{a}_{i}}{\partial \boldsymbol{x}_{i}} \|$ is bounded by $\frac{3}{2} \sqrt{n_1}G$.
\end{proposition}
\begin{proof}
\begin{equation}
\begin{aligned}
    \| h'(\boldsymbol{a}_{i}) \frac{\partial \boldsymbol{a}_{i}}{\partial \boldsymbol{x}_{i}} \|
    & = \sqrt{\sum_{i=1}^{n_1}\sum_{i=1}^{n_3}\sum_{i=1}^{n_2} (h'_{ik}w_{kj})^2}
    \\
    & = \sqrt{\sum_{i=1}^{n_1}\sum_{i=1}^{n_3}\sum_{i=1}^{n_2} (|h'_{ik}| \: |w_{kj}|)^2}
    \\
    & \leq \sqrt{\sum_{i=1}^{n_1}\sum_{i=1}^{n_3}\sum_{i=1}^{n_2} (\frac{3}{2}w_{kj})^2}
    \\
    & \leq \frac{3}{2} \sqrt{\sum_{i=1}^{n_1}G^2} = \frac{3}{2} \sqrt{n_1}G
    \\
    & \leq \frac{3}{2} \sqrt{N}G = \Psi
\end{aligned}
\end{equation}
\end{proof}
    
\begin{proposition}
\label{hp_dist}
Suppose $\Tilde{h'}(\boldsymbol{a}_{i}) \in R^{{n_1}\times{n_2}}$ and $\frac{\partial \boldsymbol{a}_{i}}{\partial \boldsymbol{x}_{i}} \in R^{{n_2}\times{n_3}}$, then $\| \Tilde{h'}(\boldsymbol{a}_{i}) \frac{\partial \boldsymbol{a}_{i}}{\partial \boldsymbol{x}_{i}} \|$ is bounded by $\sqrt{n_1}G$.
\end{proposition}
\begin{proof}
\begin{equation}
\begin{aligned}
    \| \Tilde{h'}(\boldsymbol{a}_{i}) \frac{\partial \boldsymbol{a}_{i}}{\partial \boldsymbol{x}_{i}} \|
    & = \sqrt{\sum_{i=1}^{n_1}\sum_{i=1}^{n_3}\sum_{i=1}^{n_2} (\Tilde{h'}_{ik}w_{kj})^2}
    \\
    & = \sqrt{\sum_{i=1}^{n_1}\sum_{i=1}^{n_3}\sum_{i=1}^{n_2} (|\Tilde{h'}_{ik}| \: |w_{kj}|)^2}
    \\
    & \leq \sqrt{\sum_{i=1}^{n_1}\sum_{i=1}^{n_3}\sum_{i=1}^{n_2} w_{kj}^2}
    \\
    & \leq \sqrt{\sum_{i=1}^{n_1}G^2} = \sqrt{n_1}G
    \\
    & \leq \sqrt{N}G = \Tilde{\Psi}
\end{aligned}
\end{equation}
\end{proof}

Now, we derive the error bound with respect to the weight matrix $\boldsymbol{w}_{i}$ at time step $1$, \emph{i.e.}, $\|\boldsymbol{\Tilde{w}}^{1}_i - \boldsymbol{w}^{1}_i\|$, for $i < L$.

\begin{lemma}
\label{prop_bwd}
Given a $L$-layer neural network with Hard-swish activation functions, whose backward calculation is approximated with a signed function, the backward error with respect to the weight matrix $\boldsymbol{w}_{i}$ at layer $i$ for $i 
< L$ is bounded by $\lambda G(\Psi^{L-i} + \Tilde{\Psi}^{L-i})$.
\end{lemma}

\begin{proof}
We let $\boldsymbol{g}_i$ and $\boldsymbol{\Tilde{g}}_i$ be the original gradient and the approximated counterpart propagated to the $i$th layers, such that

\begin{equation}
\begin{aligned}
& \boldsymbol{w}^{1}_i = \boldsymbol{w}^0_i - \lambda   \boldsymbol{g}_i 
\\
& \boldsymbol{g}_i= \frac{\partial L}{\partial \boldsymbol{x}_{L}} h'(\boldsymbol{a}_{L-1}) \frac{\partial \boldsymbol{a}_{L-1}}{\partial \boldsymbol{x}_{L-1}}
h'(\boldsymbol{a}_{L-2}) ... h'(\boldsymbol{a}_i)
\frac{\partial \boldsymbol{a}_{i}}{\partial \boldsymbol{w}_{i}},
\end{aligned}
\end{equation}
and
\begin{equation}
\begin{aligned}
& \boldsymbol{\Tilde{w}}^{1}_i = \boldsymbol{w}^0_i - \lambda  \boldsymbol{\Tilde{g}}_i 
\\
& \boldsymbol{\Tilde{g}}_i= \frac{\partial L}{\partial \boldsymbol{x}_{L}} 
 \Tilde{h}'(\boldsymbol{a}_{L-1}) \frac{\partial \boldsymbol{a}_{L-1}}{\partial \boldsymbol{x}_{L-1}}\Tilde{h}'(\boldsymbol{a}_{L-2}) ... \Tilde{h}'(\boldsymbol{a}_i)
\frac{\partial \boldsymbol{a}_{i}}{\partial \boldsymbol{w}_{i}} 
\text{.}
\end{aligned}
\end{equation}
From triangle inequality, Cauchy-Schwarz inequality, proposition \ref{h_dist} and \ref{hp_dist}, we have
\begin{equation}
\begin{aligned}
 &\|\boldsymbol{\Tilde{w}}^{1}_i  -  \boldsymbol{w}^{1}_i\| \\
 = &   \lambda \| \boldsymbol{g}_i 
-  \boldsymbol{\Tilde{g}}_i \| 
\\
 \leq & \lambda \| \boldsymbol{g}_i 
\| +  \lambda \| \boldsymbol{\Tilde{g}}_i \|
\\
= &\lambda  \|  \frac{\partial L}{\partial \boldsymbol{x}_{L}}  h'(\boldsymbol{a}_{L-1}) \frac{\partial \boldsymbol{a}_{L-1}}{\partial \boldsymbol{x}_{L-1}} ... h'(\boldsymbol{a}_i)
\frac{\partial \boldsymbol{a}_{i}}{\partial \boldsymbol{w}_{i}} \|
\\
\quad & + \lambda \| \frac{\partial L}{\partial \boldsymbol{x}_{L}} 
 \Tilde{h}'(\boldsymbol{a}_{L-1}) \frac{\partial \boldsymbol{a}_{L-1}}{\partial \boldsymbol{x}_{L-1}} ... \Tilde{h}'(\boldsymbol{a}_i) \frac{\partial \boldsymbol{a}_{i}}{\partial \boldsymbol{w}_{i}}  \|
\\
\leq & \lambda \|  \frac{\partial L}{\partial \boldsymbol{x}_{L}}\| 
\; \|  h'(\boldsymbol{a}_{L-1}) \frac{\partial \boldsymbol{a}_{L-1}}{\partial \boldsymbol{x}_{L-1}} \|
...
\|h'(\boldsymbol{a}_i) \frac{\partial \boldsymbol{a}_{i}}{\partial \boldsymbol{w}_{i}} \|
\\
& + \lambda \| \frac{\partial L}{\partial \boldsymbol{x}_{L}} \| \; 
 \| \Tilde{h}'(\boldsymbol{a}_{L-1})\frac{\partial \boldsymbol{a}_{L-1}}{\partial \boldsymbol{x}_{L-1}}\| 
... 
\|\Tilde{h}'(\boldsymbol{a}_i) \frac{\partial \boldsymbol{a}_{i}}{\partial \boldsymbol{w}_{i}}  \|
\\
 \leq & \lambda G (\frac{3}{2} \sqrt{N}G)^{L-i} +  \lambda G (\sqrt{N}G)^{L-i}
 \\
 = & \lambda G(\Psi^{L-i} + \Tilde{\Psi}^{L-i})
\text{.}
\end{aligned}
\end{equation}

\end{proof}

Next, we derive the error bound respect to the weight matrix at training step $T$, \emph{i.e.}, $\|\boldsymbol{\Tilde{w}}^{T}_i -  \boldsymbol{w}^{T}_i\|$ , $\boldsymbol{w}_{i}$ for $i < L$.
\begin{lemma}
\label{error_t}
Given a $L$-layer neural network with Hard-swish activation functions, whose backward calculation is approximated with a signed function, if we train the network for $T$ steps, then the error with respect to the weight matrix $\boldsymbol{w}^{T}_i$ at $L$ layer is bounded by $\lambda T G(\Psi^{L-i} + \Tilde{\Psi}^{L-i})$ at time step $T$.
\end{lemma}

\begin{proof}
From Lemma \ref{prop_bwd}, at time step $1$, we have 
\begin{equation}
\begin{aligned}
    \|\boldsymbol{\Tilde{w}}^{1}_i -  \boldsymbol{w}^{1}_i\| = \lambda \|\boldsymbol{g}^{0}_i - \boldsymbol{\Tilde{g}}^{0}_i \| 
    \leq  \lambda G(\Psi^{L-i} + \Tilde{\Psi}^{L-i})
    \text{.}
\end{aligned}
\end{equation}
We derive the error bound for $t=2$ from $t=1$ using triangle inequality:
\begin{equation}
\begin{aligned}
    \boldsymbol{w}^{2}_i = \boldsymbol{w}^{1}_i - \lambda 
    \boldsymbol{g}^1_i, \; \boldsymbol{\Tilde{w}}^{2}_i = \boldsymbol{\Tilde{w}}^{1}_i - \lambda 
    \boldsymbol{\Tilde{g}}^1_i
\text{.}
\end{aligned}
\end{equation}
Therefore, 
\begin{equation}
\begin{aligned}
    \|\boldsymbol{\Tilde{w}}^{2}_i -  \boldsymbol{w}^{2}_i\| & =  \| \boldsymbol{\Tilde{w}}^{1}_i - \lambda \boldsymbol{\Tilde{g}}^{1}_i -  \boldsymbol{w}^{1}_i + \lambda \boldsymbol{g}^1_i \|
    \\
    & =  \| \boldsymbol{\Tilde{w}}^{1}_i -  \boldsymbol{w}^{1}_i  + \lambda \boldsymbol{g}^1_i - \lambda \boldsymbol{\Tilde{g}}^{1}_i  \|
    \\
    & \leq \| \boldsymbol{\Tilde{w}}^{1}_i -  \boldsymbol{w}^{1}_i \|  + \lambda \|\boldsymbol{g}^1_i - \boldsymbol{\Tilde{g}}^{1}_i  \|
    \\
    & \leq  \lambda \|\boldsymbol{g}^{0}_i - \boldsymbol{\Tilde{g}}^{0}_i \|  +  \lambda \|\boldsymbol{g}^1_i - \boldsymbol{\Tilde{g}}^{1}_i  \|
    \\
    & \leq 2 \lambda G(\Psi^{L-i} + \Tilde{\Psi}^{L-i})
\text{.}
\end{aligned}
\end{equation}

By induction, we can conclude that the error at time step $T$ is bounded by $\lambda T G(\Psi^{L-i} + \Tilde{\Psi}^{L-i})$, \emph{i.e.},
\begin{equation}
\begin{aligned}
\|\boldsymbol{\Tilde{w}}^{T}_i -  \boldsymbol{w}^{T}_i\| 
 \leq \lambda T G(\Psi^{L-i} + \Tilde{\Psi}^{L-i})
\text{.}
\end{aligned}
\end{equation}
\end{proof}

We derive an error bound for the loss between the weights trained with MobileTL and the original weights.
We let the weights in trainable $L$ layers from MobileTL be $\boldsymbol{\Tilde{W}} = (\boldsymbol{\Tilde{w}}_1, \boldsymbol{\Tilde{w}}_2, ..., \boldsymbol{\Tilde{w}}_L)$, and the original weights are $\boldsymbol{W} = (\boldsymbol{w}_1, \boldsymbol{w}_2, ..., \boldsymbol{w}_L)$.
We assume that the network function $F(\boldsymbol{\cdot})$ is Lipschitz continuous, \emph{i.e.,} $\forall \boldsymbol{x}, \boldsymbol{y} \in \text{dom} F $, $ \|F(\boldsymbol{x}) - F(\boldsymbol{y})\| \leq M\|\boldsymbol{x}-\boldsymbol{y}\|$, and derive a bound of the loss at time step $T$, \emph{i.e.}, $\|F(\boldsymbol{\Tilde{W}}^T) - F(\boldsymbol{W}^T)\|$.

\begin{repeatthm}{th}
Given trainable $L$ layers in a neural network with Hard-swish activation functions, whose backward calculation is approximated with a signed function. If we train the $L$ layers for $T$ steps, then the loss distance between $\|F(\boldsymbol{\Tilde{W}}^T) - F(\boldsymbol{W}^T)\|$ is bounded by $\lambda M T G \left(\frac{\Psi (1-\Psi^L)}{1-\Psi} + \frac{\Tilde{\Psi} (1-\Tilde{\Psi}^L)}{1-\Tilde{\Psi}}\right)$, where $M$ is the constant from the Lipschitz continuous property of $F(\boldsymbol{\cdot})$.
\end{repeatthm}

\begin{proof}
By the assumption of Lipschitz continuous and Lemma 2, 
\begin{equation}
\begin{aligned}
    & \|F(\boldsymbol{\Tilde{W}}^T) - F(\boldsymbol{W}^T)\| 
    \\
    \leq & M\|\boldsymbol{\Tilde{W}}^T-\boldsymbol{W}^T\|
    \\
    = & M\|(\boldsymbol{\Tilde{w}}_1^T-\boldsymbol{w}_1^T) + (\boldsymbol{\Tilde{w}}^T-\boldsymbol{w}^T) + ... + (\boldsymbol{\Tilde{w}}_L^T-\boldsymbol{w}_L^T) \|
    \\
    \leq & M (\|\boldsymbol{\Tilde{w}}_1^T-\boldsymbol{w}_1^T\| + \|\boldsymbol{\Tilde{w}}^T-\boldsymbol{w}^T\| + ...  + \|\boldsymbol{\Tilde{w}}_L^T-\boldsymbol{w}_L^T\|)
    \\
    \leq & \lambda M T G\left((\Psi^{L} + ... + \Psi) + (\Tilde{\Psi}^{L} + ... + \Tilde{\Psi})\right) 
    \\
    = & \lambda M T G \left(\frac{\Psi (1-\Psi^L)}{1-\Psi} + \frac{\Tilde{\Psi} (1-\Tilde{\Psi}^L)}{1-\Tilde{\Psi}}\right)
\text{.}
\end{aligned}
\end{equation}

\end{proof}

Theorem \ref{th} states that the loss difference is bounded by the Lipschitz constant $M$, training steps $T$, and the number of approximated layers $L$. Since on-device target datasets are orders of magnitude smaller than the pre-trained datasets (necessary count $T$ is small) and we approximate the Hard-swish layers only in trainable blocks ($L$ is small), MobileTL can transfer the pre-trained weights to target datasets efficiently without incurring high accuracy drop.

\end{document}